\documentclass[conference]{IEEEtran}
\IEEEoverridecommandlockouts
\usepackage{cite}
\usepackage{amsmath,amssymb,amsfonts}
\usepackage{algorithmic}
\usepackage{graphicx}
\usepackage{textcomp}
\usepackage{xcolor}
\usepackage{subfig}
\usepackage{multirow}
\usepackage{hyperref}
\usepackage{nicefrac}
\def\BibTeX{{\rm B\kern-.05em{\sc i\kern-.025em b}\kern-.08em
    T\kern-.1667em\lower.7ex\hbox{E}\kern-.125emX}}
\begin{document}

\captionsetup[figure]{font=footnotesize, labelfont=footnotesize}

% \title{UWB Synthetic Aperture Radar Imaging for Indoor Mapping Using a Mobile Robot\\
% }
% \title{Comparative Analysis of SIFT, SURF, BRISK, AKAZE and ORB on Novel UWB SAR Imaging to Identify Loop Closures in Vision-denied Environments\\
% }

\title{Novel UWB Synthetic Aperture Radar Imaging for Mobile Robot Mapping
}
% \title{UWB SAR-based Mobile Robot Mapping with Classical Feature Descriptors: A Comparative Study}
% \title{Classical Feature Detectors on UWB SAR-based Maps for Mobile Robots: A Comparative Study}

% \title{Evaluation of Classical Feature Descriptors in UWB SAR Mapping for Mobile Robots}

% \author{\centering\IEEEauthorblockN{Charith Premachandra, Achala Athukorala, U-Xuan Tan}
\author{\centering\IEEEauthorblockN{Charith Premachandra, U-Xuan Tan}
\IEEEauthorblockA{\textit{Engineering Product Development Pillar} \\
\textit{Singapore University of Technology and Design}\\
8 Somapah Road, Singapore \\
gihan\_appuhamilage@mymail.sutd.edu.sg, uxuan\_tan@sutd.edu.sg}
% achala\_chathuranga@sutd.edu.sg
% \and
% \and
% \and
% \and
% \and
}

\maketitle

\begin{abstract}

Traditional exteroceptive sensors in mobile robots, such as LiDARs and cameras often struggle to perceive the environment in poor visibility conditions. Recently, radar technologies, such as ultra-wideband (UWB) have emerged as potential alternatives due to their ability to see through adverse environmental conditions (e.g. dust, smoke and rain). However, due to the small apertures with low directivity, the UWB radars cannot reconstruct a detailed image of its field of view (FOV) using a single scan. Hence, a virtual large aperture is synthesized by moving the radar along a mobile robot path. The resulting synthetic aperture radar (SAR) image is a high-definition representation of the surrounding environment. Hence, this paper proposes a pipeline for mobile robots to incorporate UWB radar-based SAR imaging to map an unknown environment. Finally, we evaluated the performance of classical feature detectors: SIFT, SURF, BRISK, AKAZE and ORB to identify loop closures using UWB SAR images. The experiments were conducted emulating adverse environmental conditions. The results demonstrate the viability and effectiveness of UWB SAR imaging for high-resolution environmental mapping and loop closure detection toward more robust and reliable robotic perception systems.
\end{abstract}


\section{Introduction}
Radar frequency bands are preferred over visible (e.g. camera) and near-visible bands (e.g. LiDAR) due to their ability to penetrate through adverse environmental conditions, such as smoke, dust and rain \cite{adv1}. In this context, ultra-wideband (UWB) radar exhibits excellent penetration properties attributed to its high frequency components. When it comes to obtain indoor close-range measurements, UWB radar demonstrates high Signal-to-Noise Ratio (SNR) and low power consumption compared to other radar technologies (e.g. continuous wave radar) \cite{RN83}.

% Recently, UWB radars have been incorporated in mobile robotics for Simultaneous Localization and Mapping (SLAM) in challenging environments [\#]. Those SLAM systems have been proposed to replace conventional LiDAR and Camera-based systems. The raw UWB radar observations provide the reflected waveform from the surroundings as a timeseries. The amplitudes (i.e. reflection intensities) of the raw waveform indicate the size or material of the objects in the radar's Field of View (FOV). Generally, the small aperture in UWB radar modules result in a large FOV thus affecting the spatial resolution and directionality. Hence, existing UWB radar-based SLAM systems have either extracted features (e.g. points and lines) using several reflections, or used the entire waveform as a visual template corresponding to the locations in the environment [\#\#\#]. Conversely, LiDAR and camera modules provide a feature-rich observation through a single scan. Those information are often integrated with grid maps for a better representation of the environment during SLAM. 

Recently, UWB radars have been incorporated in mobile robotics for mapping in challenging environments \cite{RN124, aoa}. Those systems have been proposed to replace conventional LiDAR and Camera-based systems. The raw UWB radar observations provide the reflected waveform from the surroundings as a timeseries. The amplitudes (i.e. reflection intensities) of the raw waveform indicate the size or material of the objects in the radar's Field of View (FOV). Generally, the small aperture in UWB radar modules result in a large FOV thus affecting the spatial resolution and directionality. Hence, existing UWB radar-based maps consist of either extracted features (e.g. points and lines) using several observations, or used the entire waveform as a visual template corresponding to the locations in the environment \cite{cp1, RN139, RN122}. Conversely, LiDAR and camera modules provide a feature-rich observation through a single scan. Those information are often integrated with grid maps for a better representation of the environment. 

\begin{figure}[t]
\centerline{\includegraphics[width=3.32in]{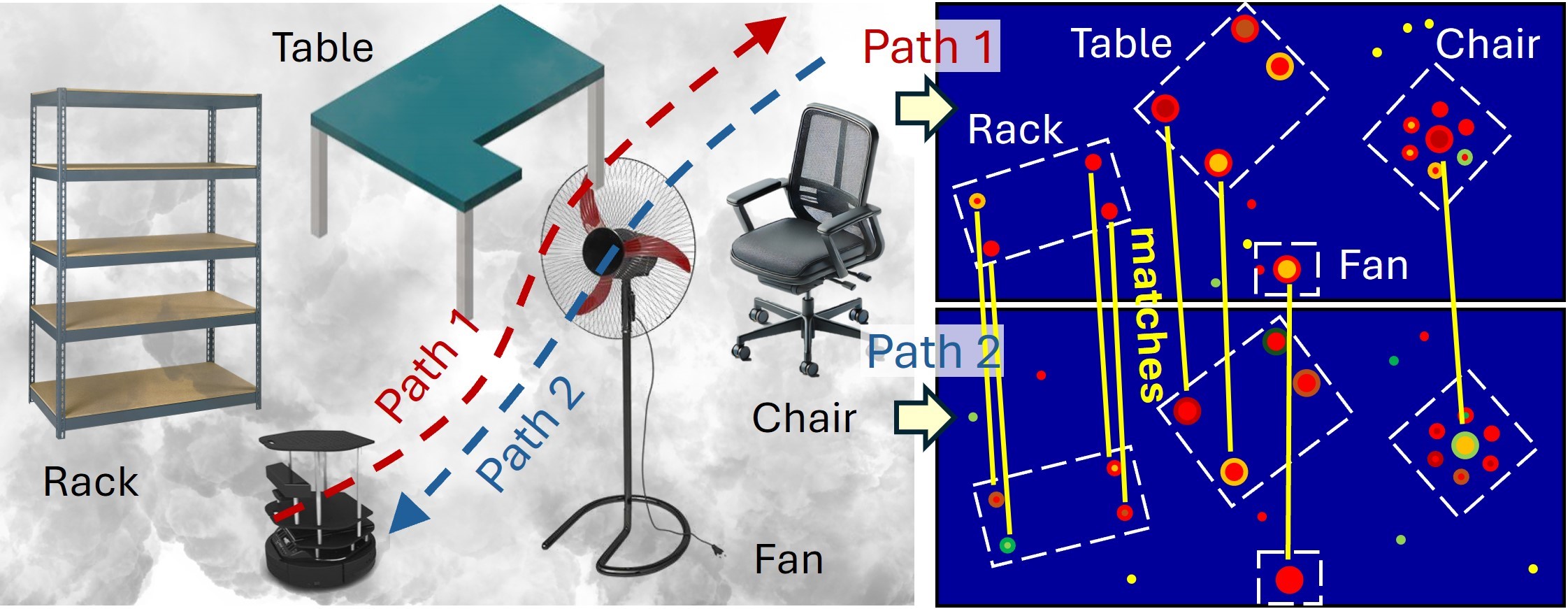}}
\caption{This paper analyses the feasibility of utilizing visual feature detectors on UWB SAR images to identify loop closures when a mobile robot explores in a vision-denied environment (e.g. smoke-filled setting).}
\label{fig_1}
\end{figure}

However, the aperture of the UWB radar can be artificially expanded by moving the radar sensor along a predefined linear \cite{lsar} or circular \cite{csar} fixed path. The radar observations are collected relative to the known poses to generate a high-resolution representation of the environment called: Synthetic Aperture Radar (SAR) image. Each pixel intensity represents the occupancy of objects in the surroundings. The features of objects with large radar cross-sections (RCS) appear brighter and vice versa.
% SAR imaging is already established in continuous wave radar systems for simultaneous localization and mapping (SLAM) \cite{sar_slam1}.
There are several algorithms to generate SAR images using UWB radar observations, such as optical algorithm \cite{bsc}, range migration algorithm \cite{rpm} and back-projection algorithm. When it comes to SAR imaging along a free path, back-projection algorithm is preferred over the others due to its flexibility in accounting for both the position and orientation of the radar system \cite{lsar, csar}. Hence, back-projection is the backbone of SAR imaging in this study. 

Meanwhile, feature extraction and matching using feature detectors have been widely utilized in the context of vision-based applications. Feature detection has been one of the fundamental components in most of the visual SLAM algorithms, especially for loop closure (e.g. VINS Mono \cite{vins_mono}, ORB-SLAM \cite{orb_slam}). Thus, several studies have analysed strengths and limitations of these feature detectors, and have suggested most versatile detector in their respective application domains \cite{comp1, comp2}. However, in contrast to SAR images, RGB camera outputs are rich in distinctive features, which facilitates more reliable feature detection. Hence, this paper evaluates the effectiveness of conventional feature detectors when applied to SAR images and examines whether their performance is consistent with that observed in RGB images.

The contributions of this paper are as follows:
\begin {itemize}
    \item We propose a complete pipeline to generate UWB SAR images using state-of-the-art UWB radar modules to perform environmental mapping;
    \item We discuss the feasibility of using feature extraction and description algorithms: SIFT, SURF, BRISK, AKAZE and ORB on SAR images to perform loop closures; 
    \item We publicly share our experiment datasets and code within the ROS2 framework to support future research.\footnote{\href{https://github.com/CPrem95/uwb_sar}{https://github.com/CPrem95/uwb\_sar}}
\end {itemize}

% Hence, similar to the LiDAR scans, SAR images allow loop closure detection via procedures such as scan-matching.

% \section{Related Work}

\begin{figure}[t]
\centerline{\includegraphics[trim=0in 0 0in 0, clip, width=3.1in]{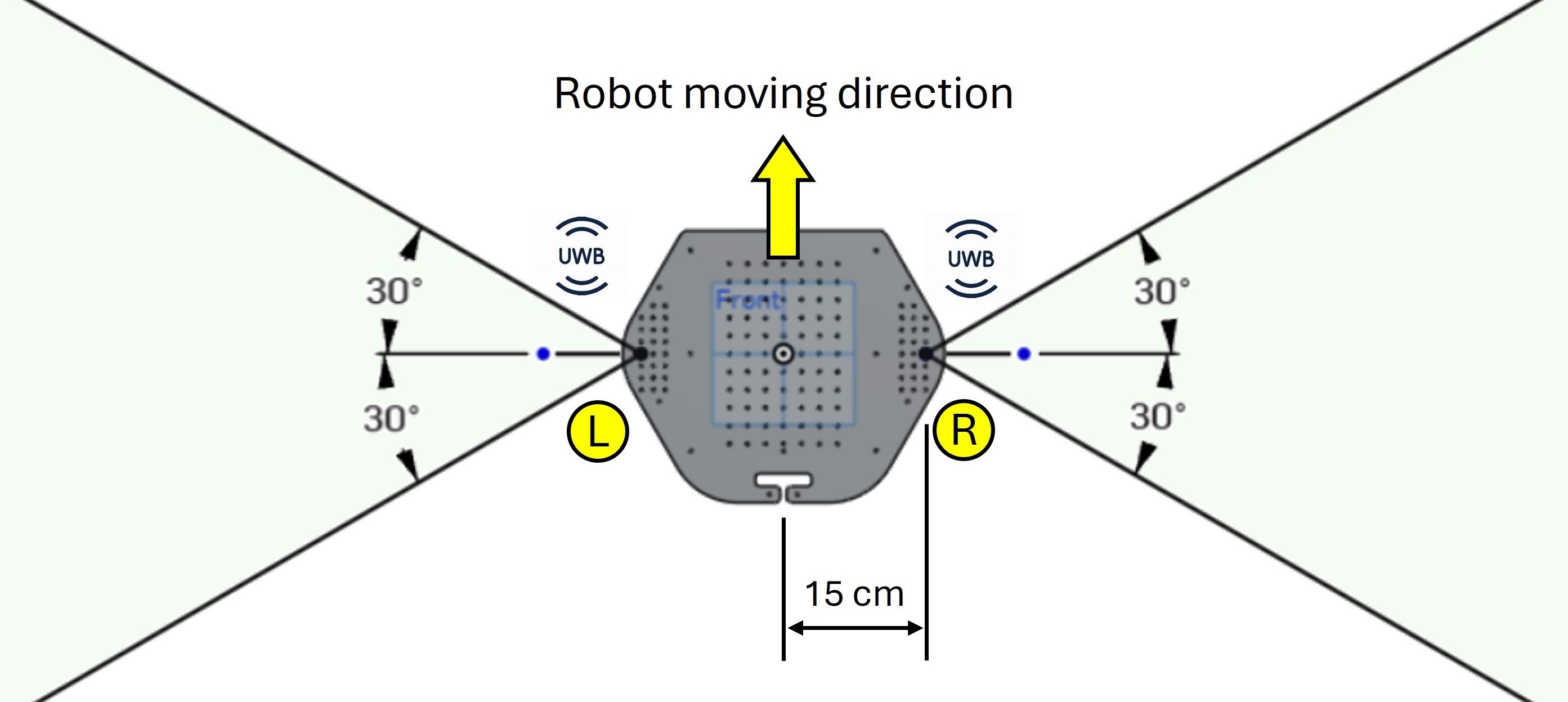}}
\caption{The plan view of the UWB radars mounted on a non-holonomic robot. The sensors are oriented perpendicular to the moving direction.}
\label{config}
\end{figure}

\begin{figure}[t]
\centerline{\includegraphics[width=2.75in]{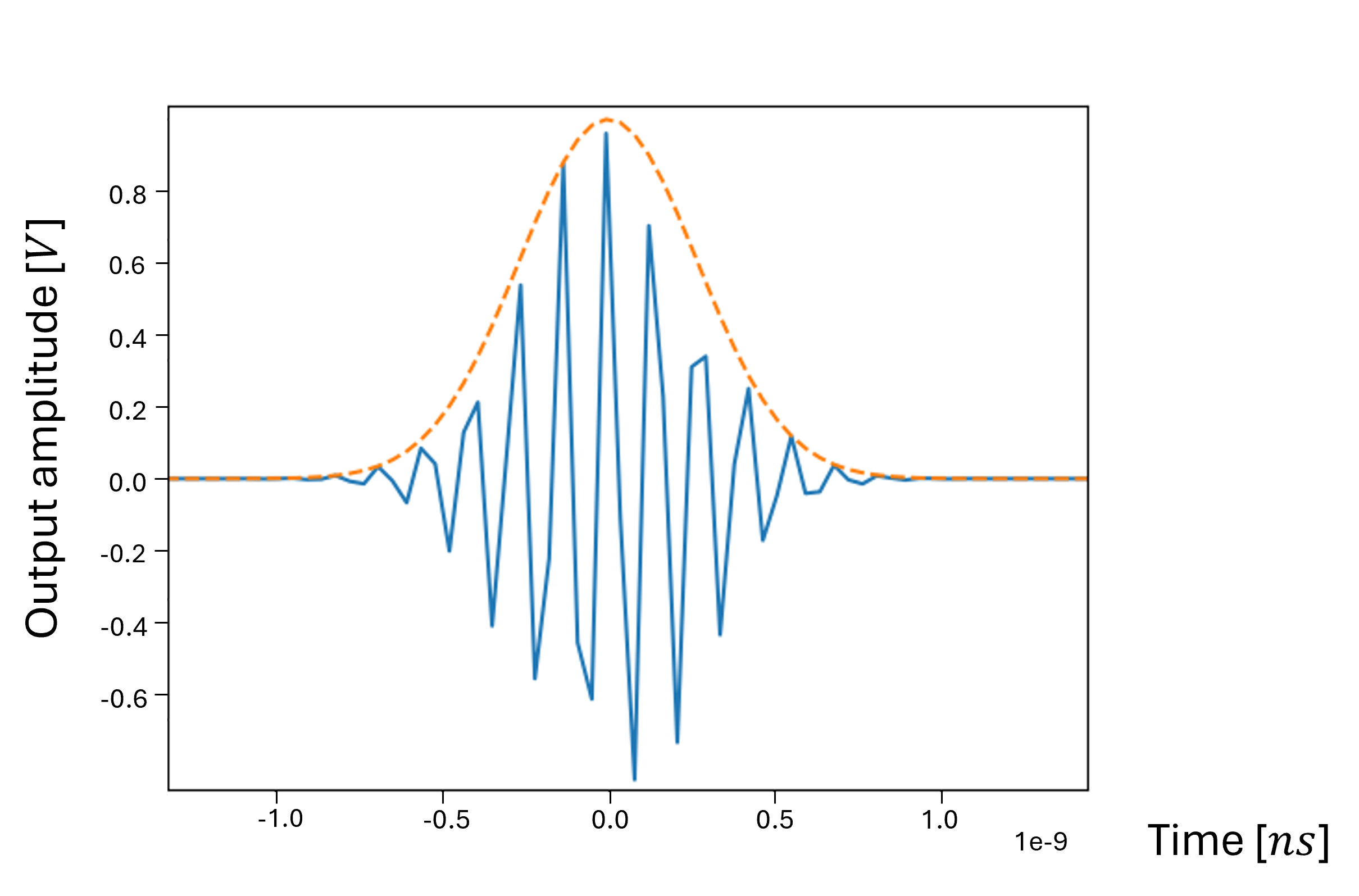}}
\caption{Transmitted Gaussian pulse. The specifications are given in Table \ref{tab1}. Recreated using \textit{scipy.signal.gausspulse}.}
\label{fig_pulse}
\end{figure} 

\section{UWB SAR Imaging}
\subsection{Sensor Configuration}
The primary sensor used in this study is the state-of-the-art LT102 UWB radar by ARIA Sensing$^\text{®}$.
Two radar modules are mounted perpendicular to the moving direction of a non-holonomic robot as in Fig. \ref{config}. We consider an effective beamwidth of 60$^\circ$ from each antenna, and operates within a range of 0.4 - 3 m.

\begin{table}[t!]
\caption{Specifications of the UWB radar}
\begin{center}
\begin{tabular}{|c|c|}
\hline
% \textbf{Table}&\multicolumn{3}{|c|}{\textbf{Table Column Head}} \\
% \cline{2-4} 
% \textbf{Head} & \textbf{\textit{Table column subhead}}& \textbf{\textit{Subhead}}& \textbf{\textit{Subhead}} \\
% \hline
Sampling frequency $f_s$& 23.328 GHz\\
\hline
Center frequency $f_c$& 7.29 GHz\\
\hline
Bandwidth& 2 GHz\\
\hline
% Pulse-repetition frequency& 14 MHz\\
% \hline
Pulse amplitude& 1.0 V\\
\hline
% \multicolumn{2}{l}{$^{\mathrm{a}}$Sample of a Table footnote.}
\end{tabular}
\label{tab1}
\end{center}
\end{table}

\begin{figure}[t]
\centerline{\includegraphics[width=3.3in]{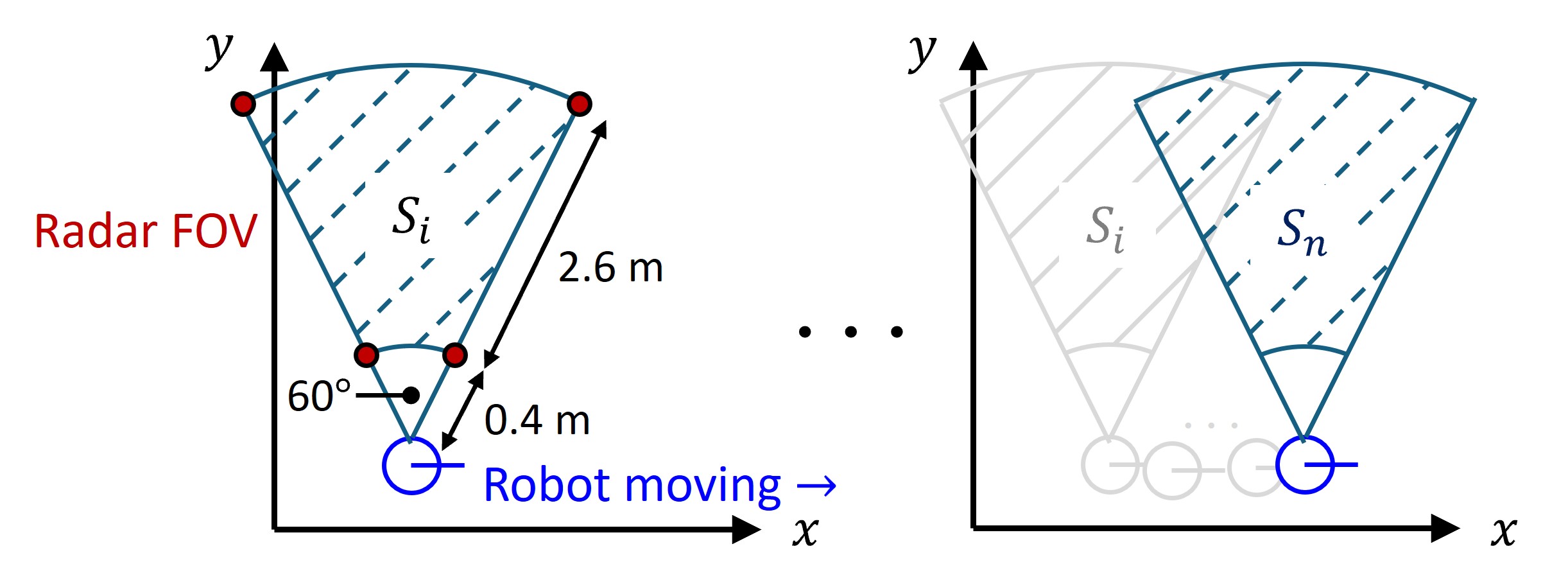}}
\caption{Overview of the back-projection algorithm. The SAR image is  reconstructed by coherently summing the received radar signals along the time-delay paths corresponding to each image pixel.}
\label{fig_bp}
\end{figure}

\subsection{Imaging Algorithm}
SAR imaging has several key components: radar observation acquisition, signal preprocessing and image reconstruction.

\subsubsection{Trajectory Estimate}
SAR imaging relies on the pose estimations of the radar system to assign observations to the corresponding pixels.
Existing systems use guide tracks with encoders or measurements from global navigation satellite system (GNSS) with inertial measurement units (IMU) to obtain pose information \cite{sar_gnss, lsar}. 
This paper proposes using wheel odometry as pose estimates over short distances to generate UWB SAR images as local views of the environment.  
% This paper proposes fusing data from the inertial measurement unit (IMU) with wheel odometry to obtain more accurate pose estimations of the robot. The robot operating system (ROS2) provides a ready-to-use package \textit{robot\_localization} for sensor fusion using the extended Kalman filter (EKF) \cite{loc2}.

\subsubsection{Range Compression}
The process of range compression is typically performed using Matched Filtering. The idea is to correlate the received signal $r(t)$ with the transmitted pulse $s(t)$, which maximizes the SNR at the correct range.
The specifications of the transmitted Gaussian pulse are given in Table \ref{tab1}. The recreated $s(t)$ is shown in Fig. \ref{fig_pulse}.
The matched filter is given by:
\begin{equation}
h(t) = s^*(t)
\end{equation}
where \( s^*(t) \) is the complex conjugate of the transmitted signal \( s(t) \). The output of the matched filter is:
\begin{equation}
   y(t) = r(t) * h(t) 
\end{equation}
where \( * \) denotes the convolution operation between the received signal \( r(t) \) and the matched filter \( h(t) \). Later, $y(t)$ is fed to the back-projection algorithm as the observation from the UWB radar.

\subsubsection{Back-projection Algorithm}
\begin{figure}[t]
\centerline{\includegraphics[width=3.25in]{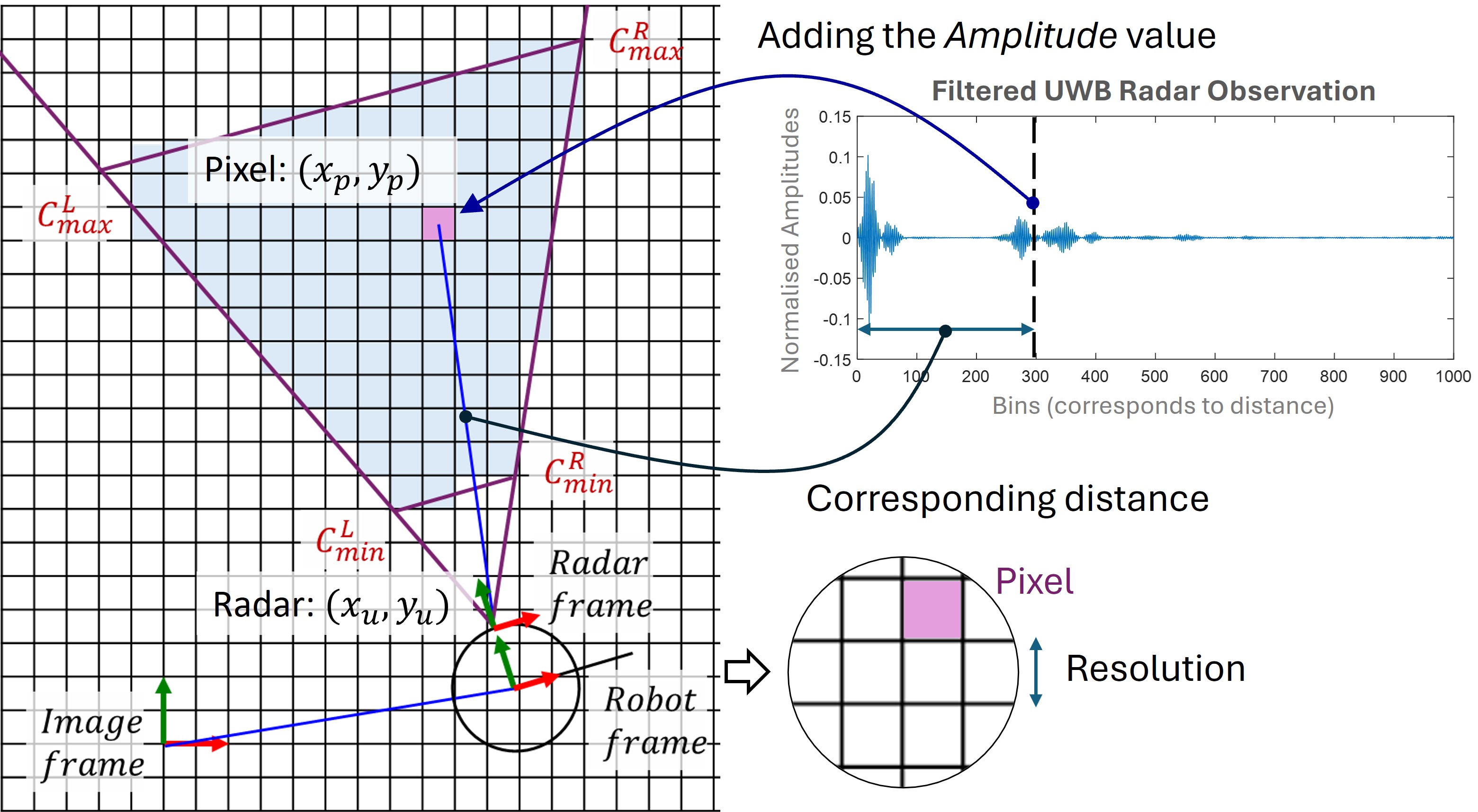}}
\caption{An illustration of the back-projection algorithm from a programming perspective. The grid (left) represents pixels of the SAR image. All coordinates are calculated w.r.t. the image coordinate frame \textbf{$I$}.}
\label{fig_pix}. 
\end{figure}

The basic idea behind `back-projecting' is to estimate the occupancies of the objects using radar scans (i.e. back-projecting the observations to the environment). Several scans are stacked from several positions to yield the final result (see Fig. \ref{fig_bp}).
The SAR image consists of pixels as illustrated in Fig. \ref{fig_pix}. Each pixel represents a specific area in the environment similar to cells in an occupancy grid. A high-definition (i.e. high resolution) SAR image represents a given area with more pixels, thereby capturing finer spatial details. The distance between a pixel $\boldsymbol{P}$ and the UWB radar $\boldsymbol{U}$ is given by:
\begin{equation}
R = \sqrt{(x_p - x_u)^2 + (y_p - y_u)^2}
\end{equation}

The corresponding Bin index (see the bin axis of filtered observation $y(t)$ in Fig. \ref{fig_pix}) can be found using:
\begin{equation}
\Delta d = \frac{c}{2.f_s}; \hspace{1em} \text{Bin index} = \frac{R}{\Delta d} \in \mathbb{Z}
\end{equation}
where $c$ is the speed of light, and refer Table \ref{tab1} for $f_s$. 

Finally, the amplitude in the $y(t)$ signal that corresponds to the Bin index is extracted for each pixel within the radar's FOV.
We can define another variable $S_i$ to denote the $i$th scan. Each $S_i$ comprises pixel values corresponding to the radar's FOV. The final $\boldsymbol{SAR}$ image is a summation of all $n$ of these $S_i$ scans (see Fig. \ref{fig_bp}). 
% Notice that the FOV depends on the effective beamwidth and range of the UWB radar. 
\begin{equation}
SAR = \sum_{i=1}^{n} S_i
\end{equation}

\begin{figure}[t]
\centerline{\includegraphics[width=3.35in]{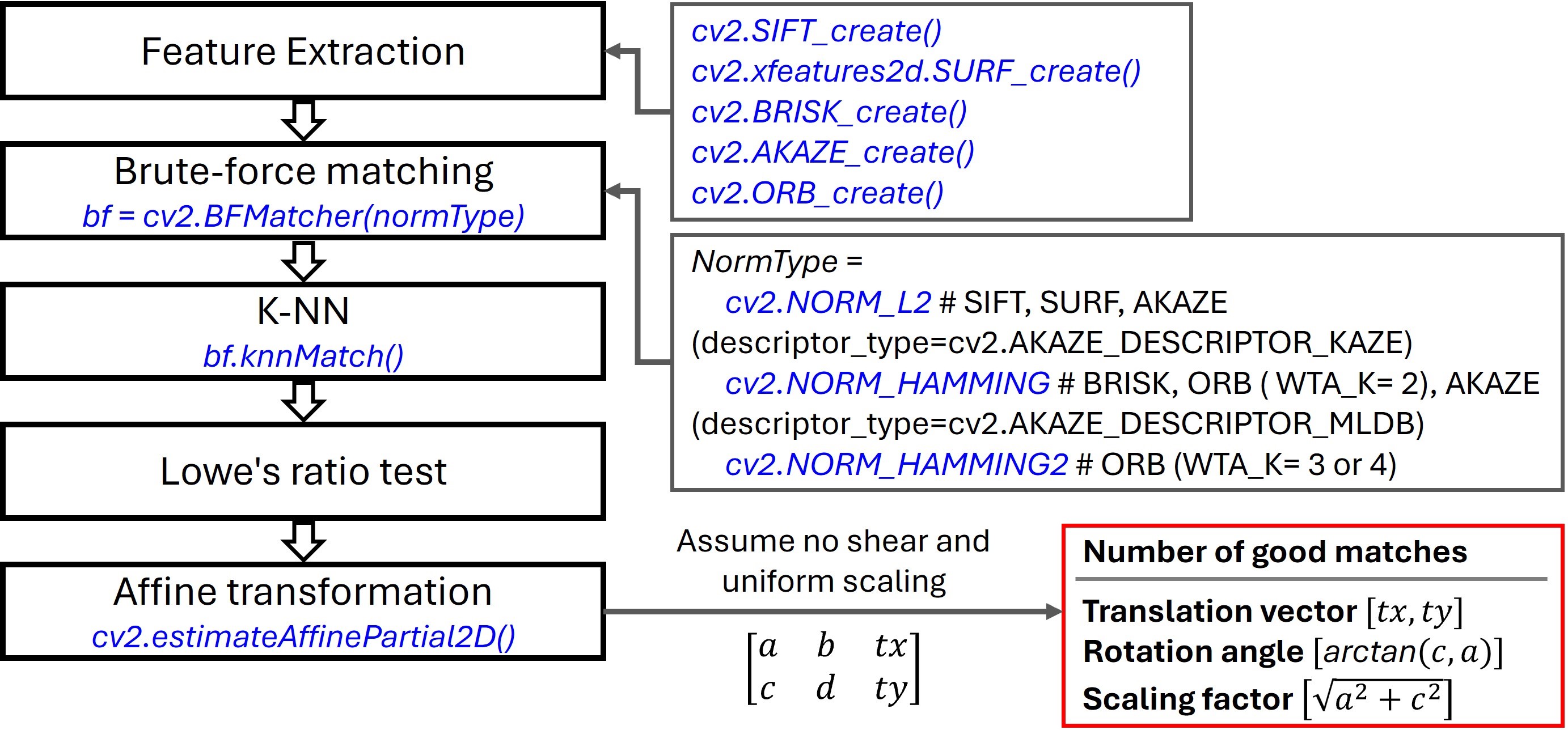}}
\caption{Feature extraction and matching pipeline to obtain the affine transformation and the number of consistent correspondences (i.e. good matches).}
\label{match_pipe}
\end{figure}

\begin{figure}[t]
\centerline{\includegraphics[width=3in]{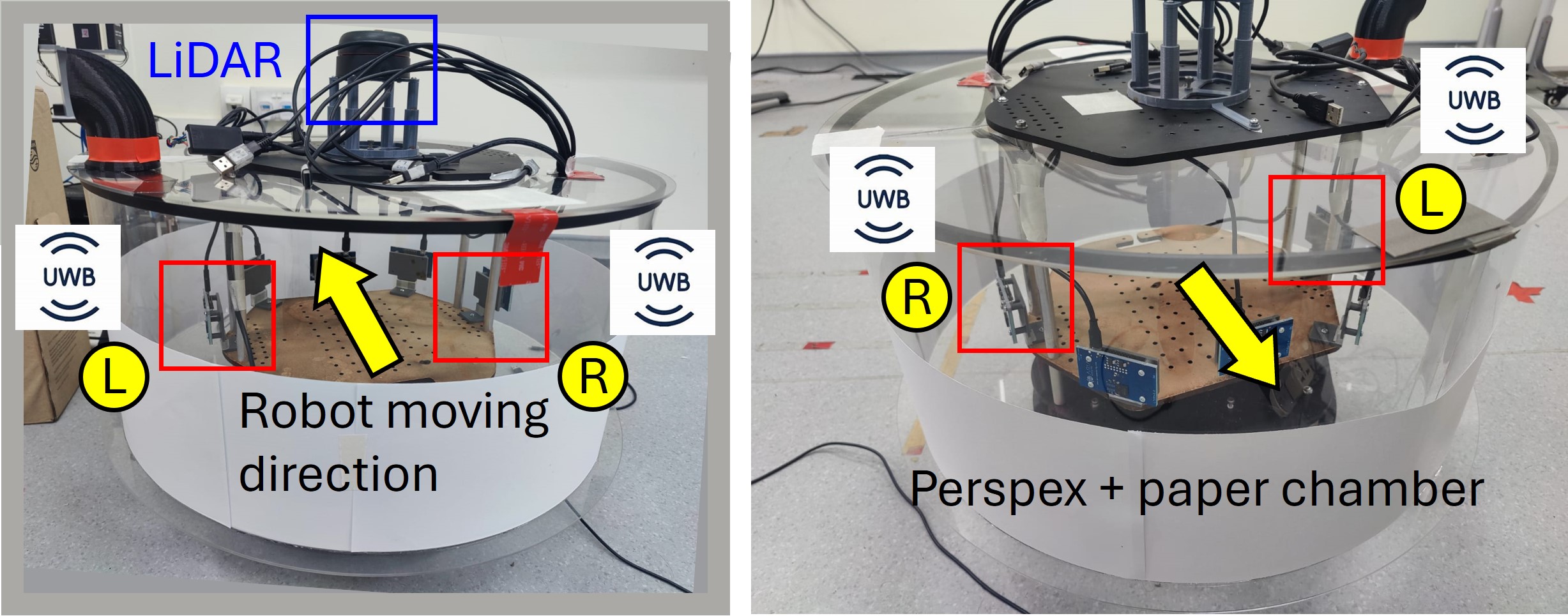}}
\caption{The mobile robot with the sensing setup. A non-holonomic
 mobile robot is used: TurtleBot2. The UWB radar modules (LT102 by ARIA Sensing) are mounted on both sides. A LiDAR is used to obtain ground truth. The robot is covered with thick perspex and papers to emulate a vision-denied scenario.}
\label{robot}
\end{figure}

\begin{figure}[t]
\centerline{\includegraphics[width=3.25in]{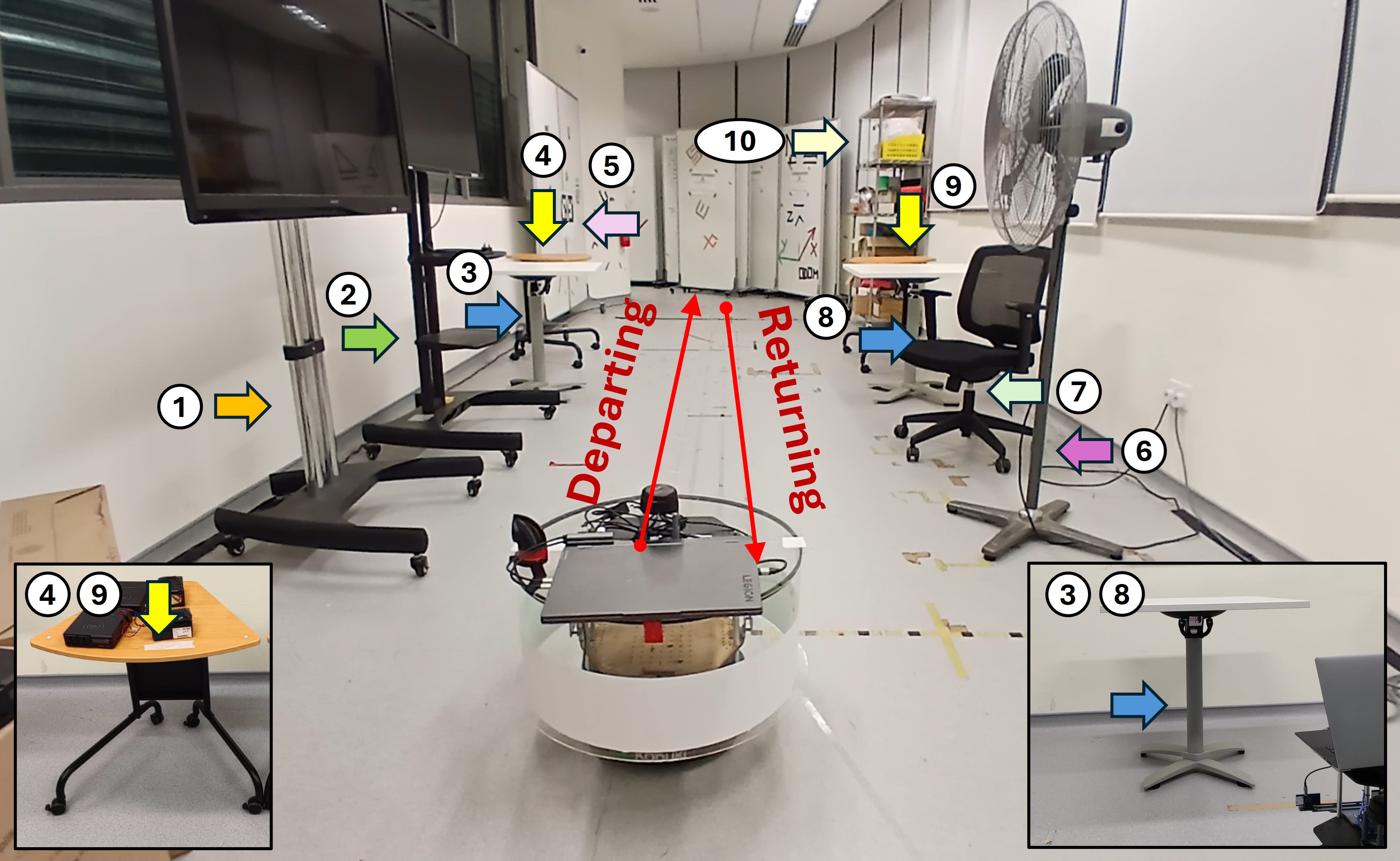}}
\caption{The indoor environment where the SAR imaging experiment was conducted. The environment consists of objects with complex shapes (i.e. features), such as office chairs, and study tables.}
\label{env}
\end{figure}

During programming, it is time-consuming to check whether each pixel lies within the FOV. 
To optimize this, OpenCV was utilized to extract only the selected region. Initially, the vertices of the FOV were computed based on the beamwidth and range. These vertices were then input into the \textit{fillPoly()} function to generate a mask that isolates the FOV. Fig. \ref{fig_pix} illustrates and summarizes the process involved in extracting pixel intensities for each $S_i$.

\section{Feature Detectors to Identify Loop Closures}

In this study, we employ five popular feature detector and descriptor algorithms \cite{vision, comp1}: Scale Invariant Feature Transform (SIFT), Speeded Up Robust Feature (SURF), Binary Robust Invariant Scalable Keypoints (BRISK), Accelerated-KAZE (AKAZE) and Oriented Fast and Rotated BRIEF (ORB) on SAR images to identify visual features. 
Subsequently, the extracted features are matched against potential candidates to facilitate loop closure detection. A candidate SAR image typically corresponds to a previously observed scene, indicating a potential revisit to a known location. 

Matching between two images involves several steps. Initially, local features (i.e. keypoints) and their corresponding descriptors are extracted. A brute-force matcher is then used to compute pairwise distances (e.g. L2 norm or Hamming distance) between the descriptors. After that, two closest matches are found by applying K-nearest neighbors. Then we use Lowe's ratio test to filter ambiguous matches. Finally, we apply RANSAC to obtain the best matching affine transformation (homography transformation is not required since the SAR images are inherently 2D without perspective distortions). Fig. \ref{match_pipe} includes the matching pipeline along with the OpenCV functions.
We claim loop closure by evaluating the estimated affine transformation matrix together with the number of consistent features (i.e. good matches). Further insight into the proposed loop closure hypothesis validation criteria can be found in Section~\ref{sec_loop}.

\section{Experiments and Results}

Fig. \ref{robot} illustrates the configuration of the sensors (i.e.
 UWB radar, and lidar) mounted on a TurtleBot2. There is an onboard computer: a laptop with an i7 processor to get UWB radar observations and lidar scans. Another laptop with an i9 processor is connected to the robot via WiFi. It executes SAR imaging within ROS2 framework from the operator’s side.
 The robot is surrounded by a Perspex + paper chamber to emulate a vision-denied scenario. 
 
\begin{figure}[t!]
\centering
	% \subfloat[Initial SAR image (left) and the positive image (right).]{\includegraphics[width=3.3in]{SAR_init.jpg}%
	% 	\label{step1}}
	% \hfil
	% \subfloat[Positive image.]{\includegraphics[height=1.75in]{positive_SAR.png}%
	% 	\label{step2}}
	% \hfil
	% \subfloat[Smoothened positive image using Gaussian blur.]{\includegraphics[height=1.7in]{SAR_proc.png}%
	% 	\label{step3}}
	% \hfil
    \centerline{\includegraphics[height=1.75in]{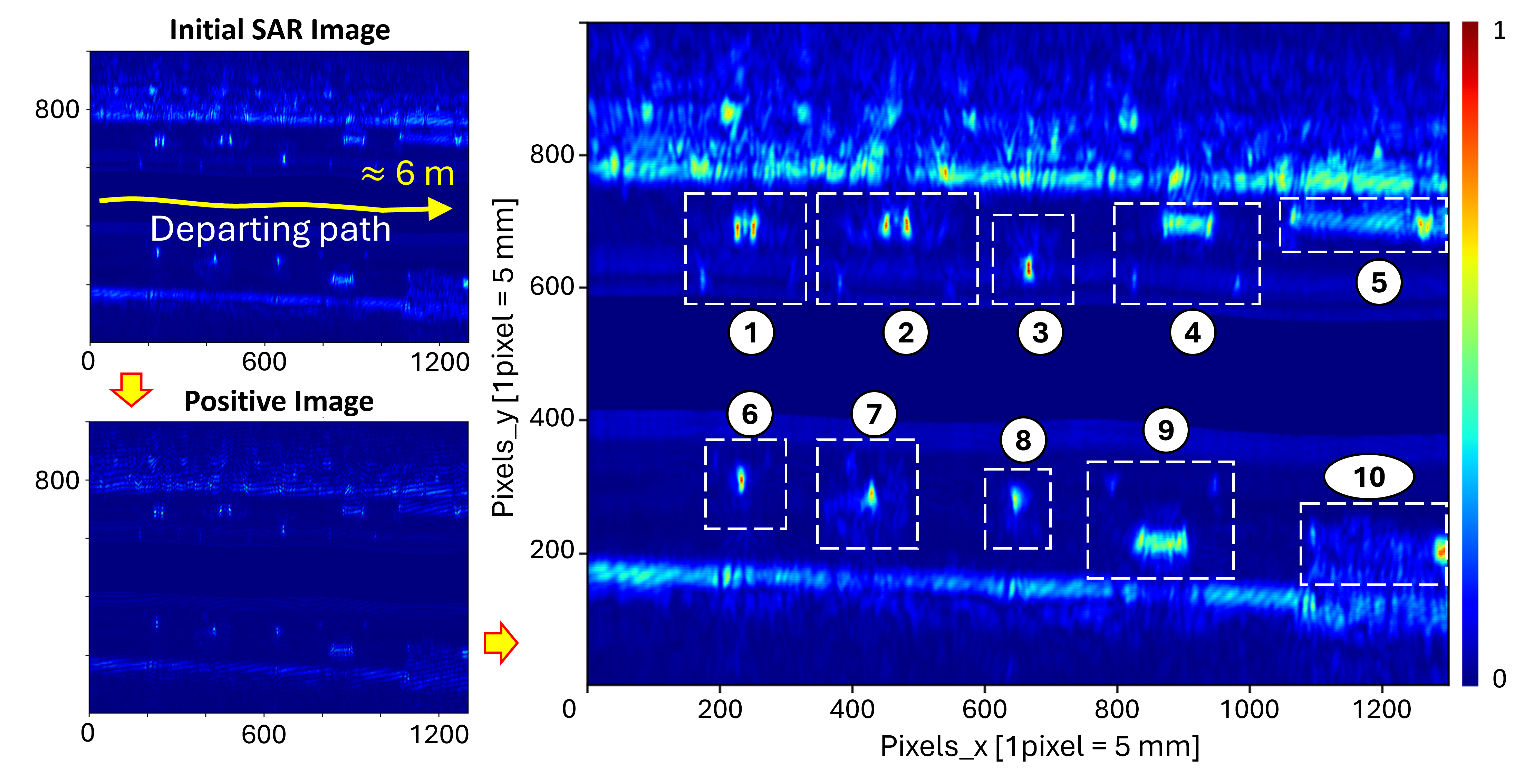}}
    \caption{SAR image post-processing steps to enhance details: initial SAR image $\rightarrow$ positive image $\rightarrow$ smoothened image (Gaussian blur). The identified objects are annotated using white boxes. This SAR image was generated along the departing path as shown in Fig. \ref{env} with a resolution of 5 mm per pixel.}
    \label{fig:sar}
\end{figure}

\begin{figure}[t]
\centerline{\includegraphics[width=3.3in]{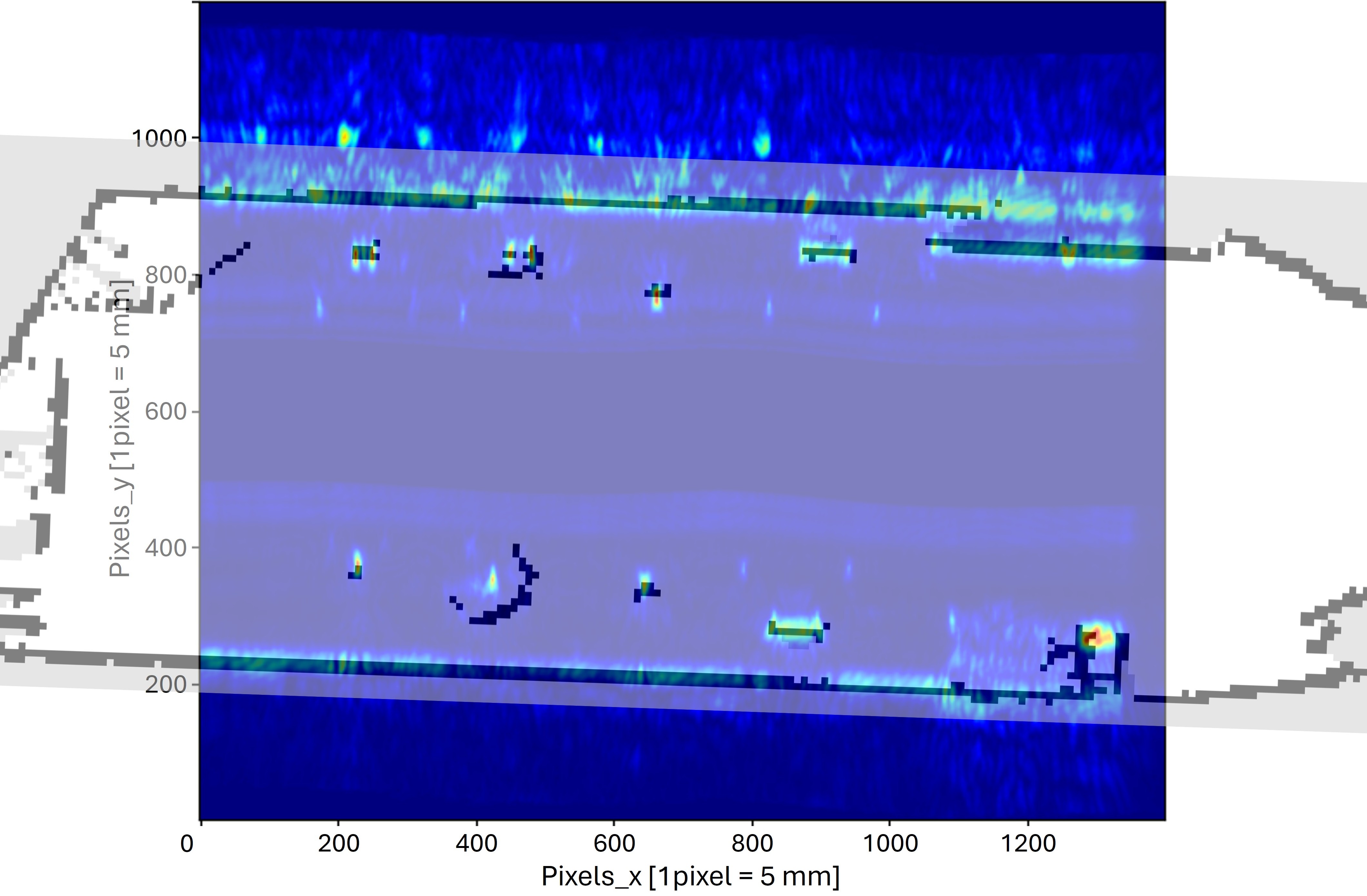}}
\caption{The ground truth occupancy grid map obtained from the LiDAR + ROS2 SLAM Toolbox is superimposed on the SAR image. The features from both representations align with each other, especially the walls. Refer the shared github repository for the cell-wise difference evaluation result.}
\label{comparison}
\end{figure}

\begin{figure}[t]
\centerline{\includegraphics[width=3.1in]{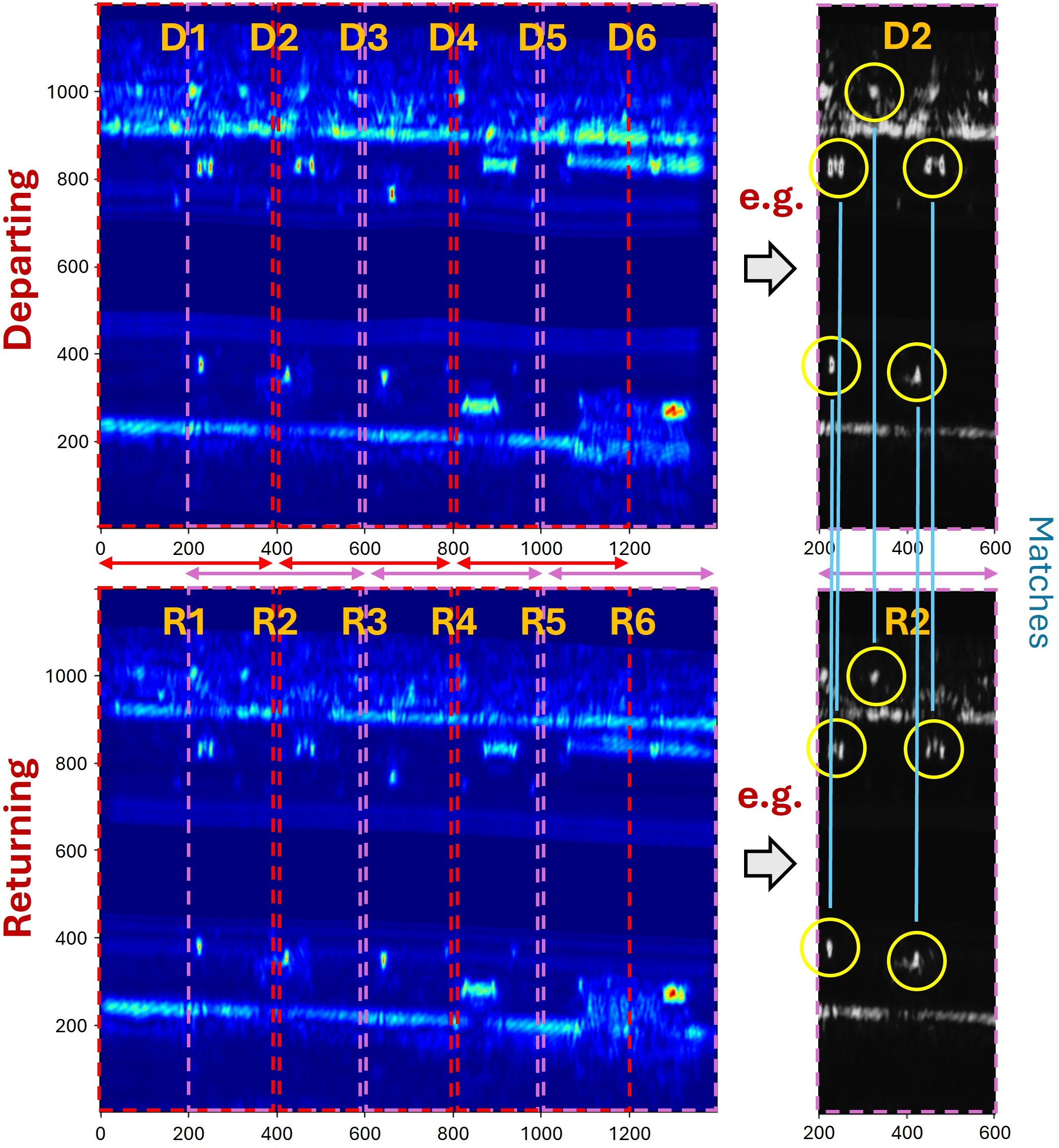}}
\caption{The generated UWB SAR images along the departing and returning paths (see Fig. \ref{env}) were split into six regions. A pairwise feature extraction and matching between these regions is conducted to evaluate the feature detectors (e.g. region D2 and R2 is a loop closure and it should match).}
\label{regions}
\end{figure}

\subsection{UWB SAR Imaging}
To evaluate the performance of UWB SAR imaging in the context of environmental mapping, we teleoperated the robot within a cluttered indoor environment as shown in Fig. \ref{env}. The generated SAR image along the departing path is shown in Fig. \ref{fig:sar}.
Initially, the features of the image are enhanced by removing the negative pixels \cite{pimg}:
\begin{equation}
\text{Positive\_Image} = Re(\text{Image}) + Abs(\text{Image})
\end{equation}

\begin{figure*}[t]
\centerline{\includegraphics[width=6.5in]{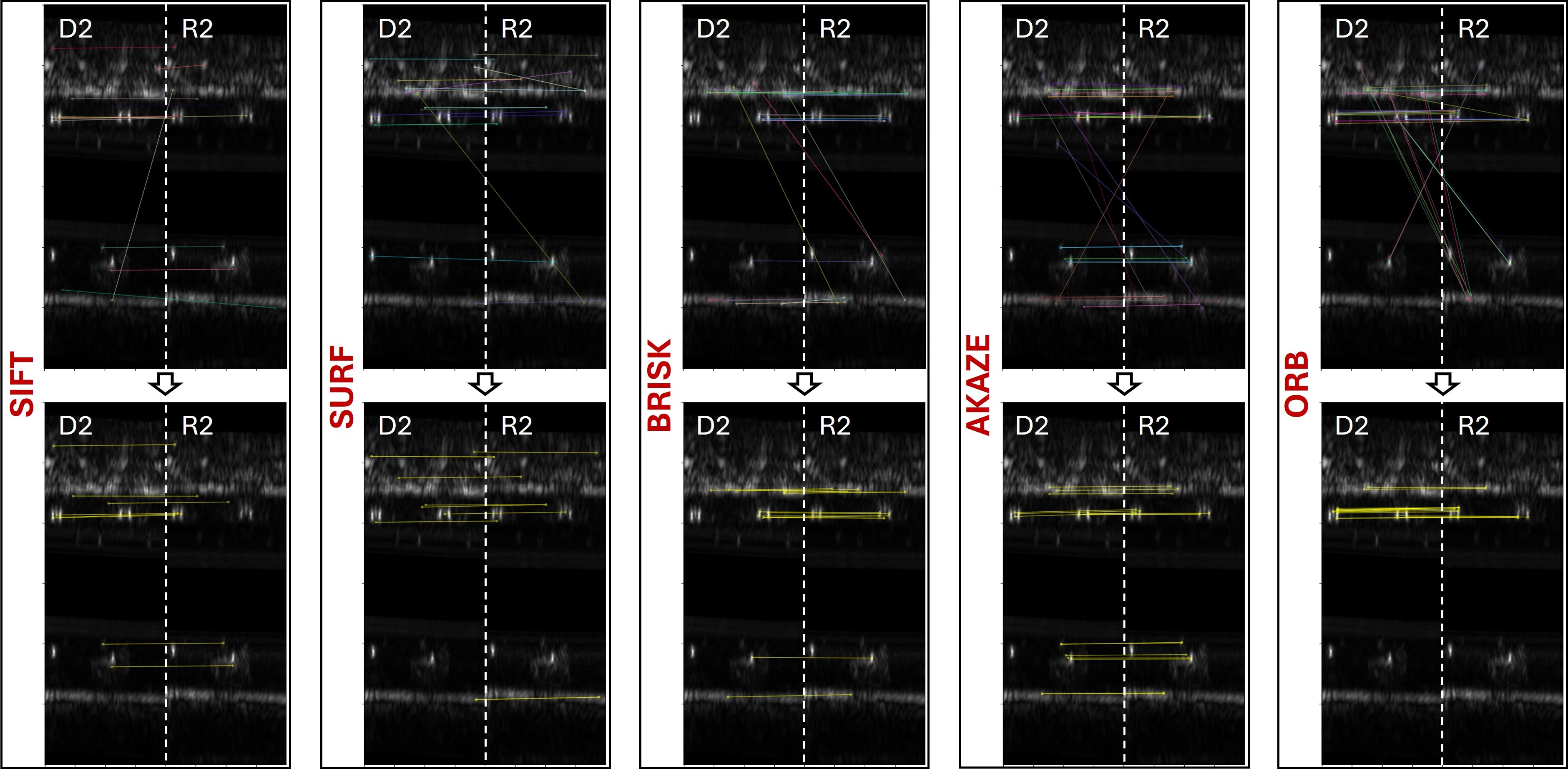}}
\caption{Feature matching example from UWB SAR images during a loop closing event: e.g. region D2 $\xrightarrow{}$ R2. Resulting matches from the Lowe's ratio test (top) and the RANSAC filtered matches (i.e. good matches - bottom). Compared to the other descriptors, AKAZE demonstrates a notable performance in detecting isolated prominent features (e.g. feature \#7 - chair in Fig. \ref{env}).} 
\label{fig: result: all detectors}
\end{figure*}

The resulting positive image is then smoothened by applying Gaussian blur to reduce noise. This process enhances the prominence of the features in the final SAR image, distinguishing them more clearly from the surroundings. We can identify the objects in the environment using the prominent features in the SAR image (e.g. object: TV stand \#1 has two metal poles, which are notable in Fig. \ref{fig:sar}). 

Meanwhile, we obtained the ground truth using the LiDAR scans + ROS2 SLAM Toolbox. The occupancy grid map from the ground truth was then superimposed onto the SAR image to assess the accuracy of object locations. As shown in Fig. \ref{comparison}, the ground truth coincides with the SAR image (cell-wise difference = \nicefrac{differing\_pixels }{ total\_pixels} = 1397 / 16800 $\approx$ 9\%). This result validates the suitability of UWB SAR images for mapping in the mobile robotics domain.

\subsection{Evaluating Feature Detectors on SAR Images}

In this experiment, we consider both UWB SAR images generated along the departing and returning paths of the robot as shown in Fig. \ref{env}.

\subsubsection{Image Preparation and Evaluation Strategy}

\begin{table}[t]
	%	\vspace{1em}
	\caption{OpenCV: modified parameters}\centering
	\begin{center}
		\begin{tabular}{|p{0.05\textwidth}|p{0.15\textwidth}|p{0.21\textwidth}|}
		\hline
		\textbf{Detector}&\textbf{modified parameter(s)}&\textbf{Remarks}\\
            \hline
            SIFT&   \textit{contrastThreshold}=0.015& Reduce to get more keypoints (KPs)\\
            \hline
            SURF&   \textit{hessianThreshold}=200 \newline \textit{nOctaveLayers}=4 & Reduce threshold and increase layers to get more KPs.\\
            \hline
            BRISK&  \textit{thresh}=15 & Increase to get more KPs.\\ 
            \hline
            AKAZE& \textit{threshold}=0.0005 \newline \textit{descriptor\_type}=\newline
            {\tiny cv2.AKAZE\_DESCRIPTOR\_KAZE} & Reduce threshold to obtain more KPs. Changing the descriptor type improved matching.\\
            \hline
            ORB&   \textit{fastThreshold}=15 \newline \textit{WTA\_K}=4 & Reduce threshold to get more KPs. Increasing \textit{WTA\_K} improved matching.\\
            \hline
		\end{tabular}
		\label{tab2}
	\end{center}
\end{table}

\begin{table}[htbp]
\caption{Feature detector and descriptor performance across loop closing events}
\centering
\scriptsize
\begin{tabular}{|p{0.023\textwidth}|p{0.04\textwidth}|p{0.054\textwidth}|p{0.04\textwidth}|p{0.05\textwidth}|p{0.14\textwidth}|}
\hline
\textbf{Desc.} & \textbf{Loop} & \textbf{\#Keypts \newline SAR\_1, 2} & \textbf{\%Good\newline matches} & \textbf{Matching time(ms)} &\textbf{Affine transformation: \newline {\tiny scale, tx \textcolor{black}{(mm)}, ty \textcolor{black}{(mm)}, rot.($^\circ$)}} \\
\hline
\multirow{6}{*}{\rotatebox{90}{SIFT}} 
     & D1→R1 & 200, 200 & 34.2 &            & 0.99, 29.8, -13.4, -0.35 \\
     & D2→R2 & 200, 200 & 53.3 & Average:   & 0.99, 21.9, -28.03, -0.33 \\
     & D3→R3 & 200, 200 & 52   & 82.77      & 0.99, 41.8, -19.1, -0.94 \\
     & D4→R4 & 200, 200 & 69.2 &            & 0.99, 25.9, -10.3, -0.59 \\
     & D5→R5 & 200, 200 & 53.6 &            & 0.99, -2.44, 5.32, 0.05 \\
     & D6→R6 & 200, 200 & 60   &            & 0.92, 10.72, 164.3, 2.08 \\
\hline
\multirow{6}{*}{\rotatebox{90}{SURF}} 
     & D1→R1 & 200, 200 & 53.3 &            & 1.00, 19.9, -33.2, -0.49 \\
     & D2→R2 & 200, 200 & 58.8 & Average:   & 0.99, 20.7, 3.73, 0.24 \\
     & D3→R3 & 200, 200 & 57.1 & 45.73      & 1.00, 40.2, -33.5, -0.99 \\
     & D4→R4 & 200, 200 & 60   &            & 1.00, 27.2, -27.5, -0.63 \\
     & D5→R5 & 200, 200 & 72.7 &            & 1.00, 23.2, -5.1, -0.37 \\
     & D6→R6 & 200, 200 & 38.1 &            & 0.98, -22.1, 53.4, 1.51 \\
\hline
\multirow{6}{*}{\rotatebox{90}{BRISK}}
      & D1→R1 & 200, 200 & 92.3 &           & 0.99, 23.8, -32.2, -0.58 \\
      & D2→R2 & 200, 200 & 59.1 & Average:  & 0.99, -1.27, 11.3, 0.18 \\
      & D3→R3 & 200, 200 & 67.7 & 31.09     & 1.00, 47.7, -49.1, -1.46 \\
      & D4→R4 & 200, 200 & 58.6 &           & 1.00, 36.9, -23.1, -0.72 \\
      & D5→R5 & 200, 200 & 63.6 &           & 0.99, 7.7, 18.3, -0.26 \\
      & D6→R6 & 200, 200 & 64.7 &           & 0.98, 2.3, 40.3, 0.19 \\
\hline
\multirow{6}{*}{\rotatebox{90}{AKAZE}}
      & D1→R1 & 200, 200 & 75.7 &           & 1.00, 25.1, -36.7, -0.63 \\
      & D2→R2 & 195, 168 & 64   & Average:  & 1.00, 15.3, -34.8, -0.28 \\
      & D3→R3 & 170, 150 & 78.5 & 52.42     & 1.00, 39.02, -30.02, -0.81 \\
      & D4→R4 & 200, 186 & 79.2 &           & 0.99, 34.1, -11.6, -0.62 \\
      & D5→R5 & 200, 200 & 80.8 &           & 0.99, 8.7, -0.78, -0.22 \\
      & D6→R6 & 200, 154 & 62.5 &           & 1.00, -28.4, -5.2, 0.22 \\
\hline
\multirow{6}{*}{\rotatebox{90}{ORB}}
      & D1→R1 & 200, 200 & 66.1 &           & 1.00, 27.3, -47.4,-0.66 \\
      & D2→R2 & 200, 198 & 47.4 & Average:  & 0.99, 53.3, -20.4, -1.51 \\
      & D3→R3 & 200, 192 & 62.9 & \textcolor{black}{6.39} & 1.01, 5.03, -31.7, -0.12 \\
      & D4→R4 & 200, 200 & 54.8 &           & 1.00, 24.5, -18.6, -0.51 \\
      & D5→R5 & 200, 200 & 42.3 &           & 1.00, 28.4, -6.36, -0.57 \\
      & D6→R6 & 200, 182 & 82.1 &           & 0.99, -6.83, 23.4, 0.18 \\
\hline
\end{tabular}
\label{tab3}
\end{table}

In practise, we accumulate UWB radar observations to generate an SAR image assuming that wheel odometry-based pose estimations remain accurate over short distances. Those SAR images are later used to identify previously visited areas to correct the accumulated odometry drift (i.e. loop closing). Hence, we split each SAR image into six regions: D$i$, R$i$; ($i = \{1,...,6\}$) as shown in Fig. \ref{regions} to evaluate feature extraction and matching performance of SIFT, SURF, BRISK, AKAZE, and ORB.

\begin{figure}[t]
\centerline{\includegraphics[width=3.4in]{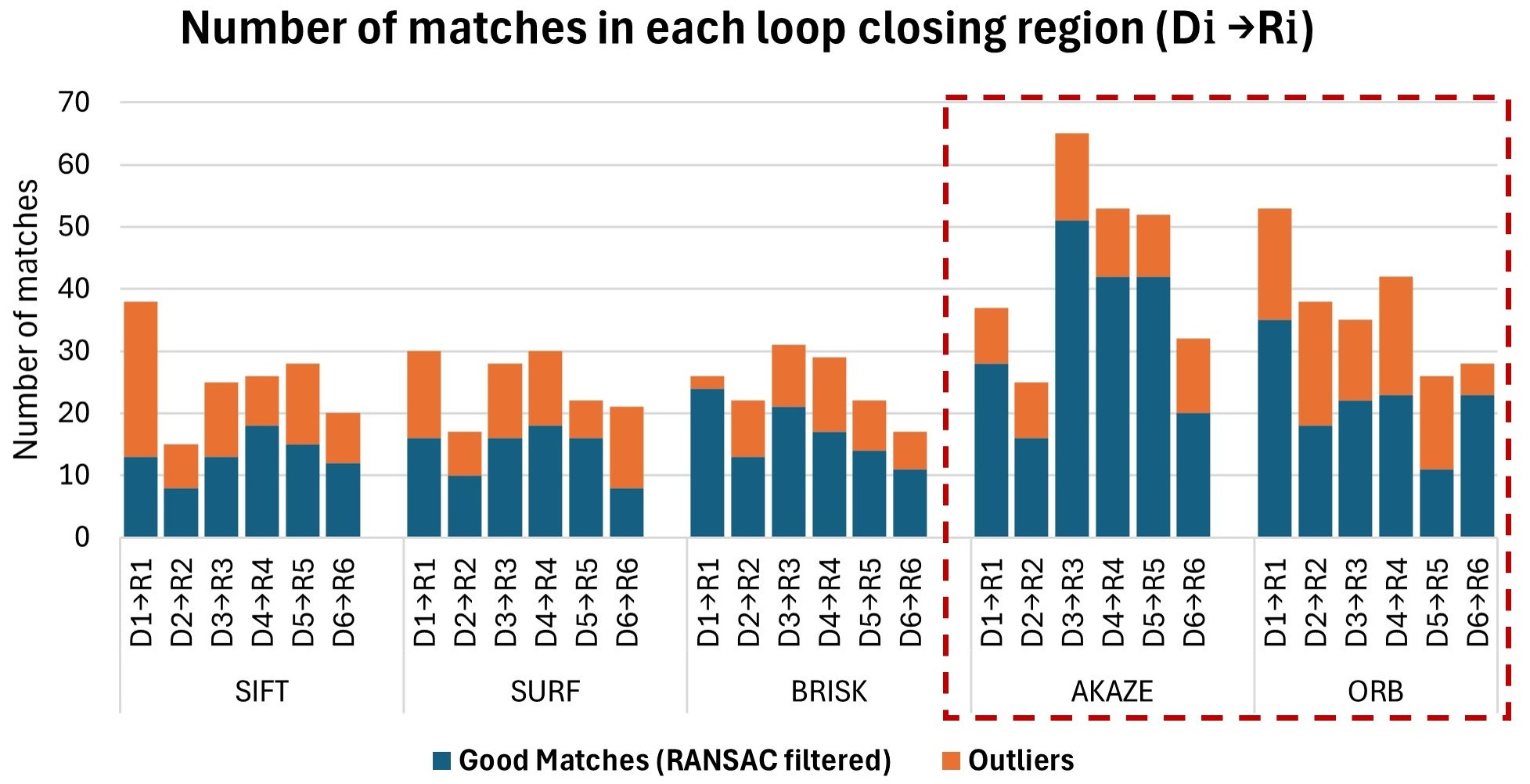}}
\caption{Evaluation of SIFT, SURF, BRISK, AKAZE and ORB detectors for UWB SAR imaging-based loop closures (D$i$, R$i$ regions are in Fig. \ref{regions})}.
\label{fig: result: all bar chart}
\end{figure}

Initially, we considered loop closing events from the SAR images: D$i$ $\rightarrow$ R$i$ where ($i = \{1,...,6\}$). The parameters of the feature detectors were tuned to obtain $\approx$ 200 best features (i.e. keypoints) from each D$i$ and R$i$ region (see Table \ref{tab2}). An intermediate result is depicted in Fig. \ref{fig: result: all detectors}, and the final results are summarised in Table \ref{tab3} and Fig. \ref{fig: result: all bar chart}. Furthermore, the consistency of the matched features was confirmed by visualizing all D$i$ $\rightarrow$ R$i$ matches (results are in the shared github repository).

According to the results, each detector was able to extract and match a distinct set of features in the SAR images, with each detector identifying a different set of features compared to the others. The filtered matches effectively captured the transformations between the two SAR images with consistent affine transformations, especially scale $\approx$ 1. However, both AKAZE and ORB stand out with more consistent matches compared to others (see Fig. \ref{fig: result: all bar chart}). 

Hence, we selected AKAZE and ORB to evaluate non-loop closing events, and the results are included in Table \ref{tab4}. Comparing Fig. \ref{fig: result: all bar chart}, Tables \ref{tab3} and \ref{tab4}, it is evident that the non-loop closing events have fewer good consistent matches with implausible affine transformations (e.g. scale $\not\approx$ 1, tx and rx are anomalously higher than expected).   

\subsubsection{Identifying Loop Closures while Exploring} \label{sec_loop} 
\begin{figure}[t]
\centerline{\includegraphics[width=3.45in]{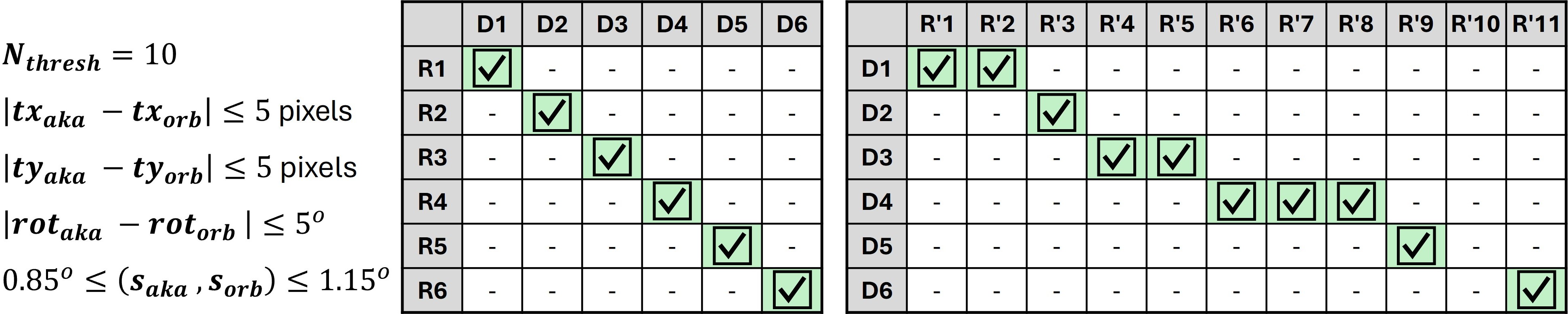}}
\caption{Evaluation of the proposed validation criteria for loop closure was conducted across all regions as in Fig. \ref{regions} (left). Additionally, R$'i$ results were obtained by splitting the returning SAR image into 11 regions (right).}
\label{fig: result: loop closures}
\end{figure}

Due to the low-feature characteristics of UWB SAR images, relying solely on either AKAZE or ORB may lead to false positive loop detections. Hence we propose considering both AKAZE and ORB features for loop closure detection during exploration. There are two validation criteria: 1) the number of good matches subjected to a threshold (i.e. [$n_{aka}$, $n_{orb}$] $\geq$ $N_{thresh}$), and; 2) consistent transformations (i.e. scales: [$s_{aka}$, $s_{orb}$] $\approx$ 1, and [$tx, ty, rot$]$_{aka} \approx$ [$tx, ty, rot$]$_{orb}$. When both criteria are satisfied, a loop closure is confirmed and the final transformation between two frames $\boldsymbol{T}$ is computed using a weighted average of the two estimates.  
\begin{equation}
  T = (n_{aka}*[tx, ty, rot]_{aka} + n_{orb}*[tx, ty, rot]_{orb})/n_{total}
\end{equation}
The proposed validation criteria were implemented on different SAR region combinations and obtained satisfactory loop detections without false positives (see Fig. \ref{fig: result: loop closures}).

\begin{table}[t]
\centering
\caption{AKAZE and ORB on non-loop closing events}
\scriptsize
\begin{tabular}{|p{0.025\textwidth}|p{0.055\textwidth}|p{0.06\textwidth}|p{0.06\textwidth}|p{0.15\textwidth}|}
\hline
\textbf{Desc.} & \textbf{non-loop \newline event} & \textbf{Total matches} & \textbf{Good matches} &\textbf{Affine transformation: \newline {\tiny scale, tx \textcolor{black}{(m)}, ty \textcolor{black}{(m)}, rot. ($^\circ$)}} \\
\hline
\multirow{6}{*}{\rotatebox{90}{AKAZE}}
        & D1→R3 & 14 & 5  & 1.16, -1.7, 0.27, 20.59 \\
        & D2→R4 & 5  & -  & - \\
        & D3→R5 & 11 & 4  & 0.92, -0.13, 0.17, -4.44 \\
        & D4→R6 & 17 & 8  & 0, 0.45, 4.5, 0.06 \\
        & D1→R4 & 8  & 4  & 1.05, 1.88, 6.67, 173.88 \\
        & D2→R5 & 9  & 3  & 0.11, 1.36, 1.59, -109.76 \\
        % & D3→R6 & 10 & 6  & 0.01, 0.45, 4.51, -21.11 \\
\hline
\multirow{6}{*}{\rotatebox{90}{ORB}}
       & D1→R3 & 23 & 7  & 1.13, -1.67, 0.39, 21.21 \\
       & D2→R4 & 25 & 8  & 0.05, 0.14, 4.13, 7.11 \\
       & D3→R5 & 20 & 10 & 0, 1.46, 4.5, 0 \\
       & D4→R6 & 19 & 7  & 0.94, 1.13, 2.50, -26.78 \\
       & D1→R4 & 25 & 6  & 1.2142, 1.14, -0.80, -25.83 \\
       & D2→R5 & 24 & 10 & 1.1361, 3.39, 6.37, -170.4 \\
       % & D3→R6 & 30 & 17 & 0.0005, 0.46, 4.50, 126 \\
\hline
\end{tabular}
\label{tab4}
\end{table}

\section{Conclusion}

This paper evaluates the feasibility of UWB SAR imaging for mapping indoor environments, and to identify loop closures using visual feature detectors: SIFT, SURF, BRISK, AKAZE and ORB. A complete pipeline is presented from SAR image generation to image enhancement, and the experimental results show that the UWB radar is capable of creating an accurate representation of the environment using SAR imaging. Although \textbf{AKAZE} is relatively poor in detecting a high number of features, it demonstrated an outstanding effectiveness in feature matching. On the other hand, \textbf{ORB} demonstrated a middle-ground between feature detection and matching accuracy, while providing a relatively high speed performance. Moreover, both detectors identify different keypoints due to the sparse feature content in SAR images. Hence, we suggest employing both \textbf{AKAZE} and \textbf{ORB} to detect loop closures by feature detection and matching. In the future, we expect to extend this approach towards UWB SAR-based SLAM.

\bibliographystyle{IEEEtran}
\bibliography{mybib}

\begin{thebibliography}{10}
\providecommand{\url}[1]{#1}
\csname url@samestyle\endcsname
\providecommand{\newblock}{\relax}
\providecommand{\bibinfo}[2]{#2}
\providecommand{\BIBentrySTDinterwordspacing}{\spaceskip=0pt\relax}
\providecommand{\BIBentryALTinterwordstretchfactor}{4}
\providecommand{\BIBentryALTinterwordspacing}{\spaceskip=\fontdimen2\font plus
\BIBentryALTinterwordstretchfactor\fontdimen3\font minus
  \fontdimen4\font\relax}
\providecommand{\BIBforeignlanguage}[2]{{%
\expandafter\ifx\csname l@#1\endcsname\relax
\typeout{** WARNING: IEEEtran.bst: No hyphenation pattern has been}%
\typeout{** loaded for the language `#1'. Using the pattern for}%
\typeout{** the default language instead.}%
\else
\language=\csname l@#1\endcsname
\fi
#2}}
\providecommand{\BIBdecl}{\relax}
\BIBdecl

\bibitem{adv1}
S.~Wang, P.~Xu, W.~Weng, L.~Niu, and R.~Wang, ``An end-to-end recognition
  method for ir-uwb radar dynamic detection mode for detecting targets in fire
  rescue scenarios,'' \emph{IEEE Internet of Things Journal}, vol.~11, no.~23,
  pp. 38\,137--38\,150, 2024.

\bibitem{RN83}
D.~Wang, S.~Yoo, and S.~H. Cho, ``Experimental comparison of ir-uwb radar and
  fmcw radar for vital signs,'' \emph{Sensors}, vol.~20, no.~22, p. 6695, 2020.

\bibitem{RN124}
W.~Chen, F.~Zhang, T.~Gu, K.~Zhou, Z.~Huo, and D.~Zhang, ``Constructing floor
  plan through smoke using ultra wideband radar,'' \emph{Proceedings of the ACM
  on Interactive, Mobile, Wearable and Ubiquitous Technologies}, vol.~5, no.~4,
  pp. 1--29, 2021.

\bibitem{aoa}
C.~Premachandra, A.~Athukorala, and U.-X. Tan, ``All-uwb slam using uwb radar
  and uwb aoa,'' \emph{IEEE Robotics and Automation Letters}, vol.~10, no.~8,
  pp. 8171--8178, 2025.

\bibitem{cp1}
H.~A. G.~C. Premachandra, R.~Liu, C.~Yuen, and U.~X. Tan, ``Uwb radar slam: an
  anchorless approach in vision denied indoor environments,'' \emph{IEEE
  Robotics and Automation Letters}, pp. 1--8, 2023.

\bibitem{RN139}
T.~Deißler and J.~Thielecke, ``Fusing odometry and sparse uwb radar
  measurements for indoor slam,'' in \emph{2013 Workshop on Sensor Data Fusion:
  Trends, Solutions, Applications (SDF)}, pp. 1--5.

\bibitem{RN122}
------, ``Feature based indoor mapping using a bat-type uwb radar,'' in
  \emph{2009 IEEE International Conference on Ultra-Wideband}.\hskip 1em plus
  0.5em minus 0.4em\relax IEEE, pp. 475--479.

\bibitem{lsar}
A.~Focsa, M.~Coca, S.-A. Toma, A.~Anghel, R.~Cacoveanu, and B.~Sebacher,
  ``Through-the-wall sar imaging based on pulson p440 off-the-shelf module,''
  in \emph{EUSAR 2024; 15th European Conference on Synthetic Aperture Radar},
  2024, pp. 566--570.

\bibitem{csar}
D.~Oloumi, P.~Boulanger, A.~Kordzadeh, and K.~Rambabu, ``Breast tumor detection
  using uwb circular-sar tomographic microwave imaging,'' in \emph{2015 37th
  Annual International Conference of the IEEE Engineering in Medicine and
  Biology Society (EMBC)}, 2015, pp. 7063--7066.

\bibitem{bsc}
\BIBentryALTinterwordspacing
N.~Cancrinus and G.~Max, ``Ultra wideband synthetic aperture radar imaging:
  Imaging algorithm,'' 2017. [Online]. Available:
  \url{https://api.semanticscholar.org/CorpusID:117390706}
\BIBentrySTDinterwordspacing

\bibitem{rpm}
S.~S. Fayazi, J.~Yang, and H.-S. Lui, ``Uwb sar imaging of near-field object
  for industrial process applications,'' in \emph{2013 7th European Conference
  on Antennas and Propagation (EuCAP)}, 2013, pp. 2245--2248.

\bibitem{vins_mono}
T.~Qin, P.~Li, and S.~Shen, ``Vins-mono: A robust and versatile monocular
  visual-inertial state estimator,'' \emph{IEEE Transactions on Robotics},
  vol.~34, no.~4, pp. 1004--1020, 2018.

\bibitem{orb_slam}
R.~Mur-Artal, J.~M.~M. Montiel, and J.~D. Tardós, ``Orb-slam: A versatile and
  accurate monocular slam system,'' \emph{IEEE Transactions on Robotics},
  vol.~31, no.~5, pp. 1147--1163, 2015.

\bibitem{comp1}
S.~A.~K. Tareen and Z.~Saleem, ``A comparative analysis of sift, surf, kaze,
  akaze, orb, and brisk,'' in \emph{2018 International Conference on Computing,
  Mathematics and Engineering Technologies (iCoMET)}, 2018, pp. 1--10.

\bibitem{comp2}
S.~A. Khan~Tareen and R.~H. Raza, ``Potential of sift, surf, kaze, akaze, orb,
  brisk, agast, and 7 more algorithms for matching extremely variant image
  pairs,'' in \emph{2023 4th International Conference on Computing, Mathematics
  and Engineering Technologies (iCoMET)}, 2023, pp. 1--6.

\bibitem{sar_gnss}
R.~Bähnemann, N.~Lawrance, L.~Streichenberg, J.~J. Chung, M.~Pantic,
  A.~Grathwohl, C.~Waldschmidt, and R.~Siegwart, ``Under the sand: Navigation
  and localization of a micro aerial vehicle for landmine detection with
  ground-penetrating synthetic aperture radar,'' \emph{Field Robotics}, vol.~2,
  pp. 1028--1067, 2022.

\bibitem{vision}
M.~Bansal, M.~Kumar, and M.~Kumar, ``2d object recognition: a comparative
  analysis of sift, surf and orb feature descriptors,'' \emph{Multimedia Tools
  and Applications}, vol.~80, no.~12, pp. 18\,839--18\,857, 2021.

\bibitem{pimg}
D.~Oloumi, M.~I. Pettersson, P.~Mousavi, and K.~Rambabu, ``Imaging of oil-well
  perforations using uwb synthetic aperture radar,'' \emph{IEEE Transactions on
  Geoscience and Remote Sensing}, vol.~53, no.~8, pp. 4510--4520, 2015.

\end{thebibliography}

\end{document}